\documentclass{article}
\usepackage{spconf,amsmath,epsfig}
\let\OLDthebibliography\thebibliography
\renewcommand\thebibliography[1]{
  \OLDthebibliography{#1}
  \setlength{\parskip}{0pt}
  \setlength{\itemsep}{0pt plus 0.3ex}
}
\usepackage{subcaption}
\usepackage[obeyspaces]{xurl}

\begin{document}\sloppy

\def\x{{\mathbf x}}
\def\L{{\cal L}}

\title{Encoding Semantic Priors into the Weights of Implicit Neural Representation}
%
\name{Zhicheng Cai,\ \   
Qiu Shen$^{\ast}$\thanks{$^{\ast}$Corresponding author: Qiu Shen (shenqiu@nju.edu.cn)\\
This work was supported in part by the National Natural Science Foundation of China under Grant 62071216 and 62231002.
}
}
%
%
\address{School of Electronic Science and Engineering, Nanjing University
}

\maketitle

\begin{abstract}
Implicit neural representation (INR) has recently emerged as a promising paradigm for signal representations, which takes coordinates as inputs and generates corresponding signal values. Since these coordinates contain no semantic features, INR fails to take any semantic information into consideration. However, semantic information has been proven critical in many vision tasks, especially for visual signal representation. This paper proposes a reparameterization method termed as SPW, which encodes the semantic priors to the weights of INR, thus making INR contain semantic information implicitly and enhancing its representational capacity. Specifically, SPW uses the Semantic Neural Network (SNN) to extract both low- and high-level semantic information of the target visual signal and generates the semantic vector, which is input into the Weight Generation Network (WGN) to generate the weights of INR model. Finally, INR uses the generated weights with semantic priors to map the coordinates to the signal values. After training, we only retain the generated weights while abandoning both SNN and WGN, thus SPW introduces no extra costs in inference. Experimental results show that SPW can improve the performance of various INR models significantly on various tasks, including image fitting, CT reconstruction, MRI reconstruction, and novel view synthesis. Further experiments illustrate that model with SPW has lower weight redundancy and learns more novel representations, validating the effectiveness of SPW.
\end{abstract}

\begin{figure*}[t]
  \centering
  \includegraphics[width=.96\textwidth]{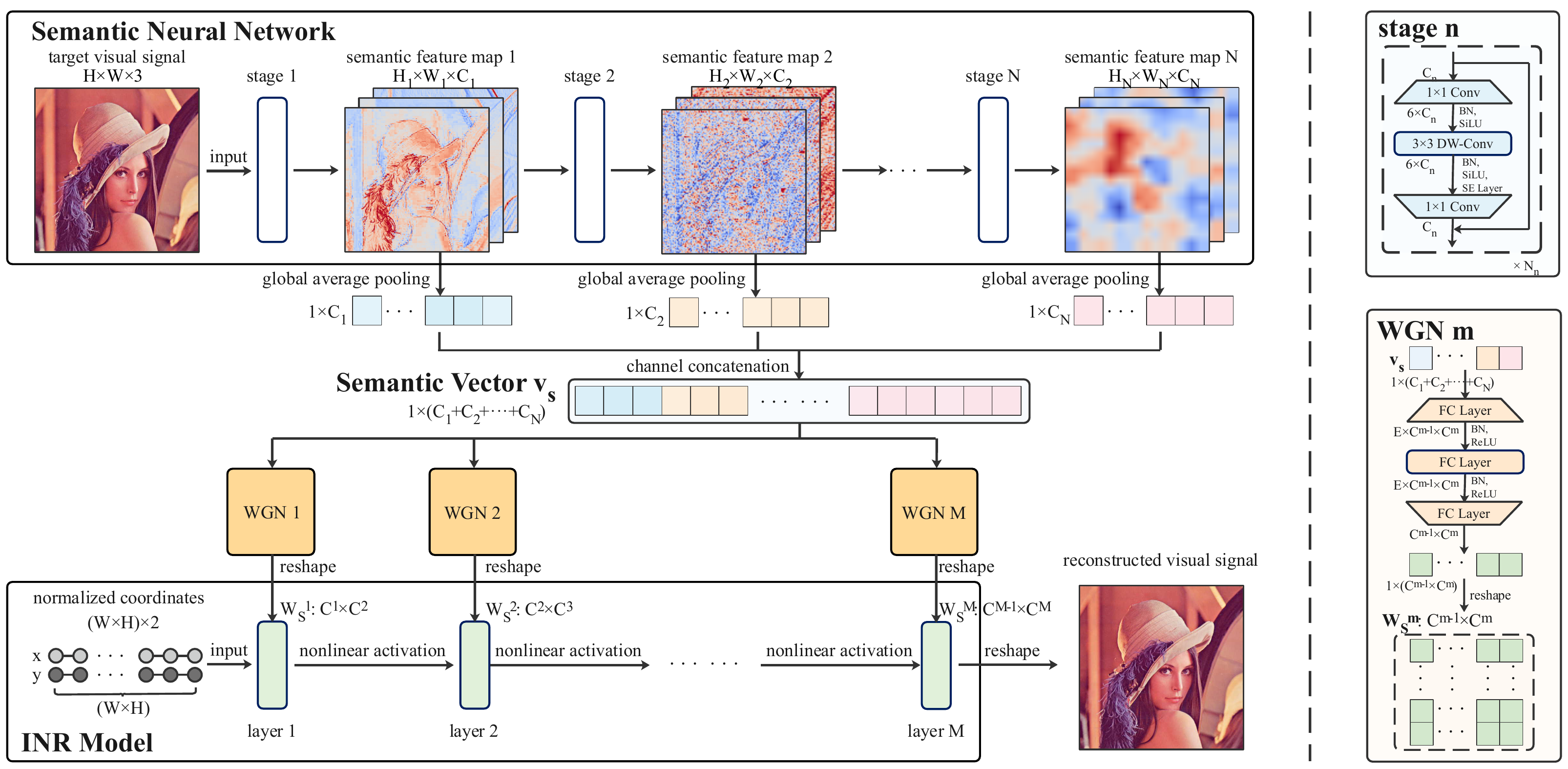} 
  \vspace{-0.5em} 
  \caption{Pipeline of SPW. The structure of each component is also exhibited.}
  \label{p1}
  \vspace{-1.5em}
\end{figure*}

\section{Introduction}
\vspace{-0.5em}
Implicit neural representation (INR) has recently emerged as a promising signal representation framework and gradually dominated in numerous vision-related tasks, including visual data representation and compression~\cite{dupont2021coin,dupont2022coin++,strumpler2022implicit, chen2021nerv, park2019deepsdf, lindell2022bacon}, scientific computing~\cite{karniadakis2021physics}, novel view synthesis~\cite{mildenhall2021nerf, pumarola2021d, barron2021mip}, and so on~\cite{tancik2020fourier,sitzmann2020implicit, saragadam2023wire, fathony2020multiplicative}. In contrast to traditional methods which learns discretized representations, INR parameterizes the continuous mapping between coordinates and corresponding signal values with a multi-layer perceptron, thus learning an efficient and compact representation of signal.

Existing INR models~\cite{tancik2020fourier,sitzmann2020implicit, saragadam2023wire, fathony2020multiplicative} only take coordinates as the input. However, coordinates contain no semantic features, thus it is hard for INR models to directly learn any semantic information. This hinders further improvement of INR representational capacity for that semantic information has been proved to be critical for many visual tasks~\cite{guo2018review,li2021survey,chen2023evolving}, particularly for learning-based signal processing~\cite{ulyanov2018deep,li2023deep} and representation~\cite{balle2018variational,cheng2020learned,minnen2018joint}. 

This paper proposes a reparameterization methodology~\cite{ding2019acnet,cai2023refconv,ma2020weightnet,cai2023falconnet,ha1609hypernetworks}, which is the first method that \textbf{encodes the semantic priors to the weights of INR} (abbreviated as SPW), thus making INR model contain semantic information implicitly and enhancing the model representational capacity. Fig.~\ref{p1} shows the overall pipeline of SPW.
To be specific, we first use the \emph{Semantic Neural Network} (SNN) to extract the semantic information of the target visual signal. This paper uses EfficientNet-B7~\cite{tan2019efficientnet} pretrained on ImageNet as the SNN, which has been proved as a powerful semantic feature extractor~\cite{tan2019efficientnet}. 
To extract both high- and low-level semantic features, we use the feature maps output by the last layer of all the stages in SNN.
We first globally average-pool these feature maps along the channel dimension and then concatenate these pooled vectors to obtain the \emph{Semantic Vector}  $\textbf{v}_{s}$. 
Subsequently, we input $\textbf{v}_{s}$ into the \emph{Weight Generation Networks} (WGNs) to generate the weights of INR. WGN is alternately composed of multiple fully-connected layers and nonlinear activation layers. Each layer of INR has its individual WGN. 
Finally, the INR model utilizes the generated weights with semantic priors $\textbf{W}_{s}$ (instead of the original weights  $\textbf{W}_{o}$ which are learned directly) to conduct the forward propagation, namely, the mapping from the coordinates to the corresponding signal value.
The parameters of WGNs are the only trainable parameters during training.
When training finished, the $\textbf{W}_{s}$ can be numerically calculated since both SNN and WGNs are fixed. Thus we only save $\textbf{W}_{s}$ and use them as the weights of INR for practical application while abandoning both SNN and WGNs.
Consequently, \textbf{SPW can enhance the representational capacity of INR models with no extra computational or memory cost for application}.

In summary, we make the following contributions,
\begin{enumerate}
\vspace{-0.7em}
\item We propose SPW, the first reparameterization methodology that encodes the semantic priors to the weights of INR to enhance the representational capacity.
\vspace{-0.7em}
\item As a general method, SPW can be applied to various INR models. We test the effectiveness of SPW on four widely-used INR models (\textit{e.g.}, PE-MLP~\cite{tancik2020fourier}, SIREN~\cite{sitzmann2020implicit}, MFN~\cite{fathony2020multiplicative}, and WIRE~\cite{saragadam2023wire}) on four tasks, including image fitting and compression, computed tomography reconstruction, magnetic resonance imaging reconstruction, and novel view synthesis with neural radiant field~\cite{mildenhall2021nerf}. 
Experimental results show that SPW can significantly improve the performance of original INR models by a clear margin (\emph{up to 1.4dB PSNR on Kodak dataset for image fitting task}). To be emphasized, the improvement is obtained without any additional computational or memory costs.
\vspace{-0.7em}
\item Further experiments show that the weights with semantic priors have lower redundancy and larger weight entropy. 
In addition, SPW makes model learn more novel representations. Both explain the effectiveness of SPW.
\end{enumerate}

\vspace{-1em}
\section{Background}
\vspace{-0.5em}
Firstly, let's review the mathematical formulation of INR~\cite{tancik2020fourier,sitzmann2020implicit, saragadam2023wire, fathony2020multiplicative,dupont2021coin,dupont2022coin++}.
For a given visual signal, for example, a color image $\textbf{I}\in\mathbf{R}^{W\times H\times3}$, $x,y\!\in\!\mathbf{R}$ are the pixel coordinates in the normalized range $[-1,1]$, $I(x,y)$ denotes the pixel values at the coordinates $x,y$.
Typically, INR is a MLP $f_{\theta}$ with L layers parameterized by $\theta = [\textbf{W}^1,\textbf{W}^2,\cdot\cdot\cdot,\textbf{W}^L]$. $W^l$ is the weights of the $l\!-\!th$ fully-connected layer and updated by gradient descent directly. 
The target of INR is to learn the mapping from the coordinates to the corresponding value, thus the optimization problem can be expressed as:
\vspace{-0.5em}
\begin{equation}
\vspace{-0.5em}
    \mathop{\arg\min}\limits_{\theta} \sum\limits_{\tilde{x},\tilde{y}}\left|f_{\theta}(\tilde{x},\tilde{y})-I(\tilde{x},\tilde{y})\right|^2_2
\end{equation}

However, since coordinates contain no semantic features, it is hard for INR models to directly learn any semantic information, thus hindering the further improvement of the representational capacity of INR models.

\section{Proposed Method}
\vspace{-0.5em}
To solve the above issue, this paper proposes SPW which encodes the semantic priors to the weights of INR.
\vspace{-1em}
\subsection{Generate the Semantic Vector}
\vspace{-0.5em}
Convolutional neural network has been proven as a powerful tool to effectively extract the visual semantic features~\cite{guo2018review,li2021survey,chen2023evolving}.
Thus we first use a convolutional neural network to extract the semantic features of the visual target signal, termed as \emph{Semantic Neural Network} (SNN).
We choose EfficientNet-B7~\cite{tan2019efficientnet} pretrained on ImageNet as the SNN.
EfficientNet-B7 comprises 7 stages (stage $n$ contains $N_n$ \emph{mobile bottleneck convolution blocks}, as shown in the upper right corner of Fig.~\ref{p1}) and 168 convolution layers in total, demonstrating superior semantic capacity, which achieves a top 1 accuracy of up to 84.4\% on the ImageNet dataset.
Since signal restoration with INR is a point-level task, thus the semantic information contained in the early stages are also significant~\cite{guo2018review}. As a consequence, we take both high- and low-level semantic information into consideration.
To be specific, we use the feature maps output by the final layer of each stage.
We first globally average-pool these feature maps $\!\in\!\mathbf {R}^{H_n\!\times\! W_n\!\times\!C_n} (n\!=\!1,2,\cdot\cdot\cdot,7)$ along the channel dimension and concatenate the pooled feature vectors $\!\in\!\mathbf{R}^{1\!\times\!C_n}$ as shown in Fig.~\ref{p1}.
Consequently, we obtain the \emph{Semantic Vector}  $\textbf{v}_{s}\!\in\!\mathbf{R}^{1\times (C_1+C_2+\cdot\cdot\cdot+C_7)}$ which contains both high- and low-level semantic information of the target visual signal.

\vspace{-1em}
\subsection{Generate the Weights with Semantic Priors}
\vspace{-0.5em}
Subsequently, we input the $\textbf{v}_{s}$ into the \emph{Weight Generation Networks} (WGNs) to generate the weights of INR. To enlarge the parameter space, each layer of INR has an independent WGN. 
As shown in the lower right corner of Fig.~\ref{p1}, WGN is composed of three fully-connected layers with an inverted residual bottleneck structure as a common practice in modern neural networks~\cite{tan2019efficientnet}.
Suppose the original weights $W^m_o\in \mathbf{R}^{C_{m-1}\times C_{m}}$ is required by the m-th INR layer. 
For the corresponding WGN $m$, the $1st$ and $2rd$ layers both have $E\!\times\!C_{m-1}\!\times\!C_{m}$ output channels, which $E$ is the expansion factor and set as 6 by default, and the final layer has $C_{m-1}\!\times\!C_{m}$ output channels. 
Finally, the WGN $m$ outputs a $1\!\times\!(C_{m-1}\!\times\!C_{m})$ vector, which is reshaped into a $C_{m-1}\!\times\!C_{m}$ matrix $\textbf{W}^m_{s}$ and then utilized as the weights of corresponding INR layer $m$ as shown in Fig.~\ref{p1}.
Consequently, the INR model utilizes the generated weights with semantic priors $\textbf{W}_{s}$ (instead of the original weights  $\textbf{W}_{o}$ that learn directly) to conduct the forward propagation, namely, the mapping from the coordinates to the corresponding signal value.
In this way, the INR model implicitly takes the semantic information of the target visual signal into consideration, thus enhancing the representational capacity.

\vspace{-1em}
\subsection{Training and Practical Application}
\vspace{-0.5em}
Since the SNN is fixed, thus we can obtain the Semantic Vector $\textbf{v}_{s}$ once at the beginning of training. During training, we only maintain $\textbf{v}_{s}$ fixed while ignoring the SNN, thus saving the training costs.
The parameters of WGNs are the only trainable parameters, which are updated by back-propagating the gradient according to the pipeline of SPW.
And the generated semantic weights $\textbf{W}_{s}$ are updated according to the forward propagation of the WGNs.
When training finished, $\textbf{W}_{s}$ can be numerically calculated since both SNN ($\textbf{v}_{s}$) and WGNs are fixed. Thus we only save $\textbf{W}_{s}$ and use them as the weights of INR for practical application, while abandoning both SNN and WGNs.
Consequently, SPW can enhance the representational capacity of the INR model without introducing any extra computational and memory cost for application.
As a general method, SPW can be applied to various INR models and improve their performance significantly, which will be validated in the experiments section.




\begin{table}[t]
\centering
\begin{tabular}{l|cccc}
\hline
Model        & 2D Image & 2D CT & 3D MRI & 5D Nerf \\
\hline
SIREN        & 25.52    & 28.30 & 26.04  & 25.44 \\
\textbf{SPW SIREN}    & \textbf{26.61}    & \textbf{29.14} & \textbf{26.82}  & \textbf{25.86} \\
\hline
PE-MLP       & 23.16    & 28.11 & 30.17  & 30.99 \\
\textbf{SPW PE-MLP}   & \textbf{24.06}   & \textbf{29.25} & \textbf{30.99}  & \textbf{31.52} \\
\hline
MFN          & 25.25    & 27.97 & 27.24  & 31.04 \\
\textbf{SPW MFN}      & \textbf{26.13}    & \textbf{28.92} & \textbf{27.71}  & \textbf{31.47} \\
\hline
WIRE         & 25.05    & 28.26 & 25.31  & 25.76 \\
\textbf{SPW WIRE}     & \textbf{25.74}    & \textbf{28.96} & \textbf{25.94}  & \textbf{26.15} \\
\hline
\end{tabular}
\caption{Experimental results of SIREN, SPW SIREN, PE-MLP and SPW PE-MLP on four tasks.}
\label{t1}
\vspace{-1em}
\end{table}

\begin{figure}[t]
  \centering
  \includegraphics[width=.48\textwidth]{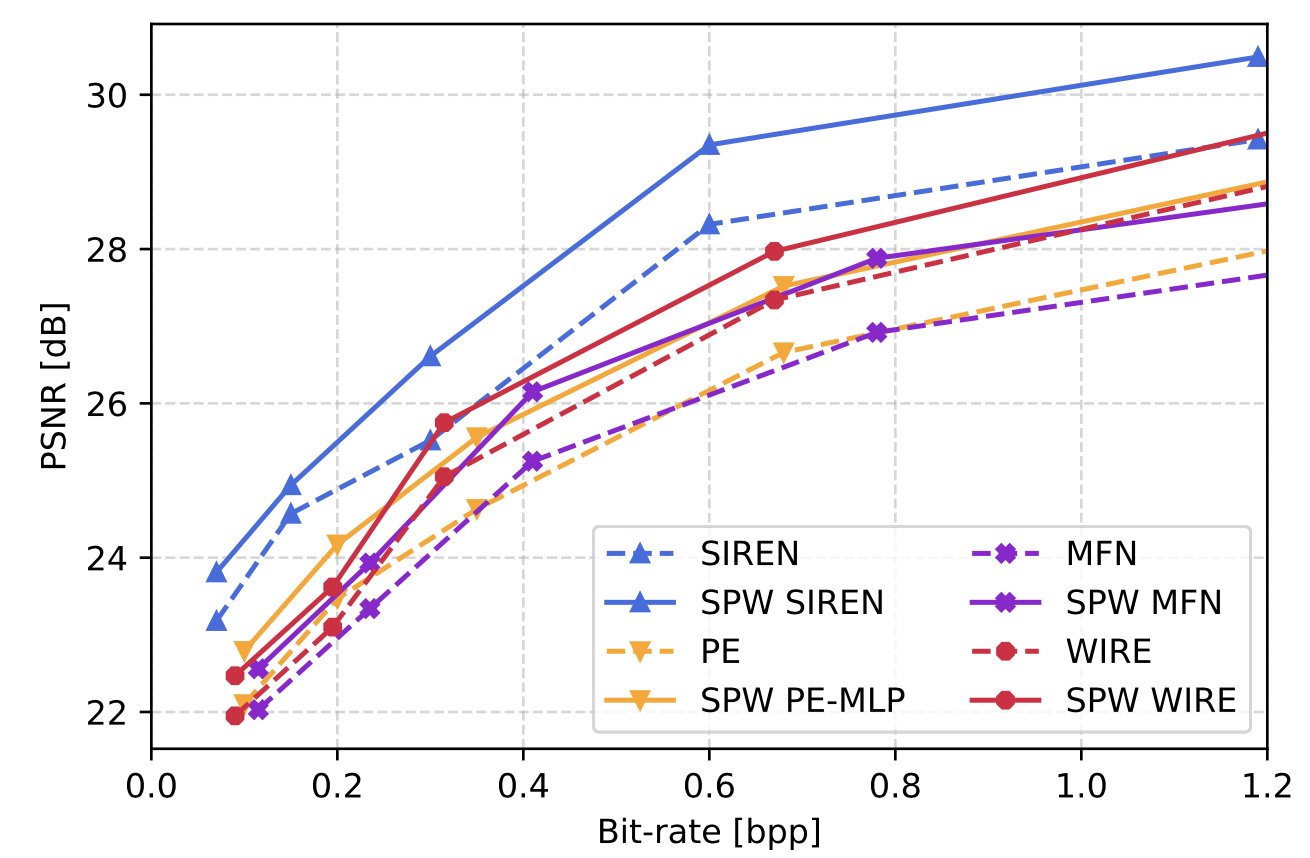} 
  \vspace{-2em} 
  \caption{Rate distortion plots of various INR models w./w.o. SPW under different bpps trained on the Kodak dataset.}
  \label{rd}
  \vspace{-1em}
\end{figure}

\vspace{-1em}
\section{Experiments}
\vspace{-0.5em}
We validate the effectiveness of SPW on four separate tasks, \textit{i.e.}, 2D image fitting and compression, 2D computed tomography (CT) reconstruction, 3D magnetic resonance imaging (MRI) reconstruction, and 5D novel view synthesis.
For these tasks, we apply SPW on four INR models, namely, PE-MLP~\cite{tancik2020fourier},  SIREN~\cite{sitzmann2020implicit}, MFN~\cite{fathony2020multiplicative} and WIRE~\cite{saragadam2023wire}, and compare their original counterparts.
As a common practice, the encoding Fourier bases of PE-MLP is set as $10$~\cite{mildenhall2021nerf}, the frequency parameter $\omega_0$  of SIREN is set as $30$~\cite{sitzmann2020implicit}, the frequency parameter $\omega$ and the spread parameter $s$ of WIRE are respectively set as $20$ and $10$~\cite{saragadam2023wire}.
For the SNN of SPW, we use EfficientNet-B7 pretrained on ImageNet, and we initialize the weights of WGNs with LeCun random initialization~\cite{lecun2002efficient}.

\vspace{-1em}
\subsection{Image Fitting}
\vspace{-0.5em}
We first use an 2D image representation task to evaluate the performance of applying SPW to INR models. 
We perform experiments on the Kodak dataset~\cite{dupont2021coin} consisting of 24 RGB images with a high resolution of $768\times512$.
To utterly explore the representational capacity of various methods, we use networks with limited parameters as introduced in \cite{dupont2021coin}, namely, networks with the architectures of (in the format of [$hidden\  layers \times hidden\  features$]) [5$\times$20], [5$\times$30], [10$\times$28], [10$\times$40], and [13$\times$49]. 
Thus it can be regarded as the image compression task using INR.
All the models are trained for 50,000 iterations using Adam optimizer~\cite{kingma2014adam} with an initial learning rate of $2e\!-\!4$.
Table.~\ref{t1} shows the average experimental results of models with [10$\times$28] measured in PSNR (Peak Signal-to-Noise Ratio). We further plot the rate-distortion curves in Fig.~\ref{rd} (each dot represents a model architecture).
As shown in Fig.~\ref{rd}, INR models with SPW achieve higher PSNR (on average up to 1dB higher) compared to their counterparts under each bit-rates burden.
The experimental results validate that through encoding the semantic priors to the weights, SPW can enhance the representational capacity of the INR model with no extra application cost.

\vspace{-1em}
\subsection{CT Reconstruction.}
\vspace{-0.5em}
In 2D CT task, we observe integral projections of a density field instead of direct supervisions. 
We train a network that takes in 2D pixel coordinates and predicts the corresponding volume density at each location.
We conduct the experiments on the x-ray colorectal dataset~\cite{saragadam2023wire,clark2013cancer}, each image has a resolution of $512\times 512$ and is emulated with 100 CT measurements.
We use networks with the architecture configuration of [2$\times$256].
All the models are trained for 20,000 iterations using Adam optimizer with a initial learning rate of $5e\!-\!3$.
Table.~\ref{t1} provides the experimental results measured in PSNR. As can be observed, INR with SPW consistently achieves higher PSNR compared to their original counterparts. 

\vspace{-1em}
\subsection{MRI Reconstruction.}
\vspace{-0.5em}
For the 3D MRI task, we observe measurements which are the Fourier transform coefficients of the atomic response to radio waves under a magnetic field. 
We train an MLP that takes in 3D voxel coordinates and predicts the corresponding intensity at each location with an indirect supervision. 
We conduct experiments on the ATLAS brain dataset~\cite{tancik2020fourier}, each sample has a volume resolution of $96^3$.
We use networks of with the architecture of [2$\times$256]. 
All the models are trained for 1,000 iterations using Adam optimizer with an initial learning rate $2e\!-\!3$. 
As shown in Table.~\ref{t1}, INR with SPW consistently obtain better performance than their original counterparts. 

\vspace{-1em}
\subsection{Novel View Synthesis.}
\vspace{-0.5em}
For the 5D NeRF experiments, we demonstrate the improvements of SPW on novel view synthesis using the neural radiance fields (NeRF)~\cite{mildenhall2021nerf}. NeRF models the 3D world as a 5D radiance fields using coordinate networks, where the input contains the 3D position and 2D viewing direction of a point and the output attributes include the RGB color and point density.
We follow the model architecture and training configuration in \cite{mildenhall2021nerf}, and conduct the experiments on the NeRF dataset~\cite{mildenhall2021nerf} (resolution of $800\times800$). 
As can be observed in Table.~\ref{t1}, SPW significantly enhances the performance of all the INR models, further validating the effectiveness of SPW.

\begin{figure}[t]
  \centering
  \includegraphics[width=.48\textwidth]{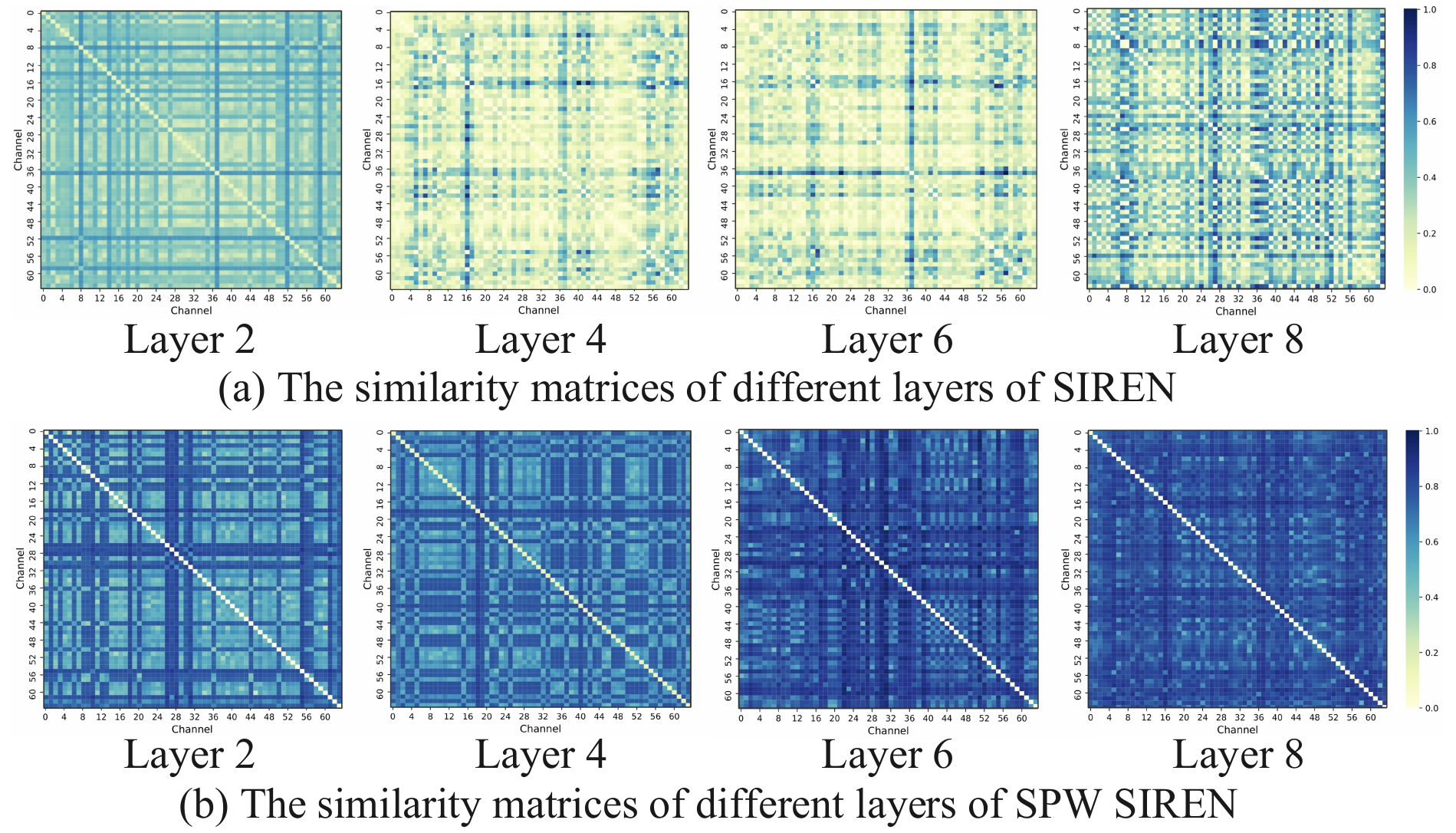} 
  \vspace{-2em} 
 \caption{The similarity matrices of different layers of SIREN and SPW-SIREN trained on Kodak. A point with a darker color represents a larger value of KL divergence, hence a lower similarity. SPW SIREN has lower similarity compared to SIREN, indicating lower weight redundancy.}
 \label{wss}
 \vspace{-1em}
\end{figure}

\begin{figure}[t]
  \centering
  \includegraphics[width=.48\textwidth]{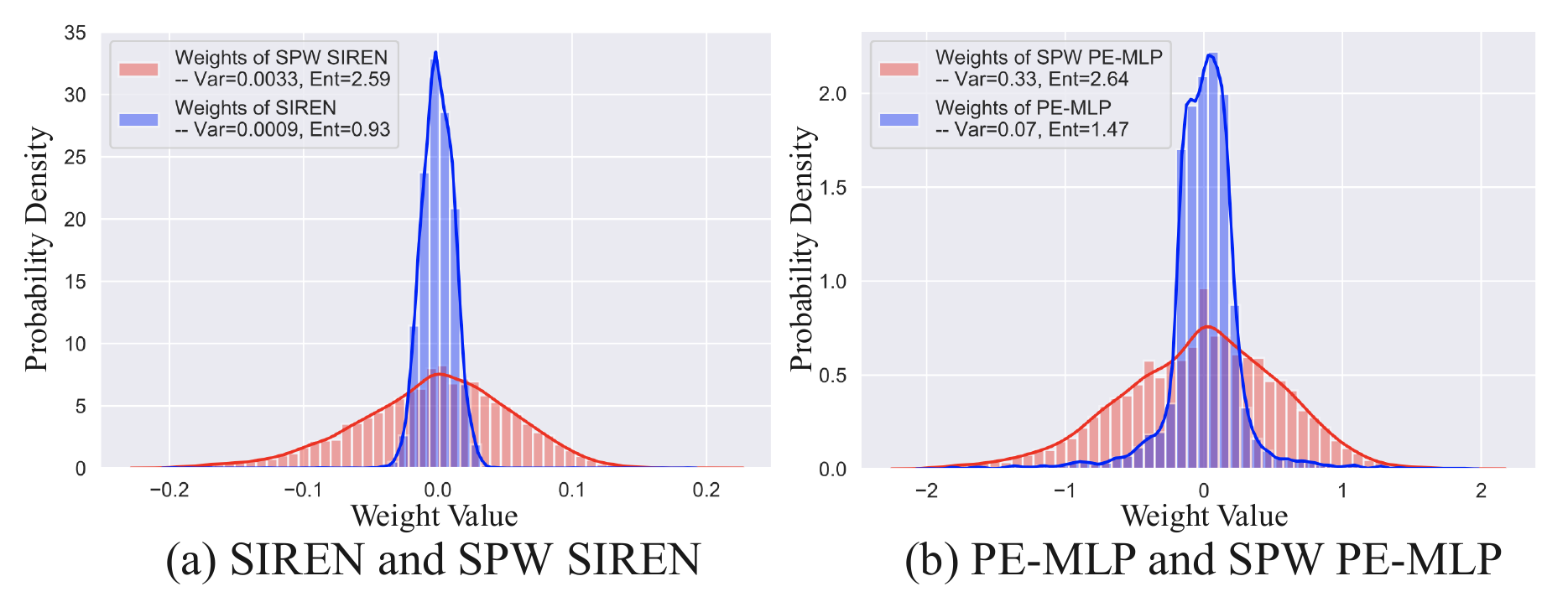} 
  \vspace{-2em} 
 \caption{The weight distribution of SIREN, SPW SIREN, PE-MLP and SPW PE-MLP trained on Kodak dataset. The INR models with SPW have larger weight entropy compared to their original counterparts.}
 \label{we}
 \vspace{-1em}
\end{figure}

\vspace{-0.5em}
\subsection{Visualization}
\vspace{-0.5em}
\textbf{Lower Weight Redundancy.} 
To explore the difference between the semantic weights and the original weights, we first compare the channel redundancy of them. As a common practice~\cite{wang2021tied}, we utilize the Kullback-Leibler (KL) divergence to measure the similarity between different pairs of channels of certain layer. \emph{A larger KL divergence indicates lower similarity, hence a lower degree of channel redundancy.} 
We calculate the $64\!\times\!64$ self-similarity matrices of (SPW) SIREN models with 8 layers and 64 channels trained on Kodak dataset.
As can be observed in Fig.~\ref{wss}, high self-similarity occurs in the original weights, while the semantic weights exhibit a lower channel redundancy.
This phenomenon is attributed to that through encoding the semantic priors to the weights, the model intends to learn diverse novel representations. Consequently, the channel redundancy is reduced and the representation diversity is enhanced, resulting in a better representational capacity~\cite{wang2021tied}. This explains the effectiveness of SPW.

\vspace{0.5em}
\noindent\textbf{Larger Weight Entropy.} 
Furthermore, we visualize the distributions of the weights of (SPW) SIREN and (SPW) PE-MLP trained on the Kodak dataset. Fig.~\ref{we} shows the results.
As can be observed, the original weights of SIREN and PE-MLP are centered around the zero, possessing a lower variance and smaller entropy. While for the weights with semantic priors, the distribution is more even and decentralized to a wider range, possessing a higher variance and larger entropy. This is consistent with the phenomenon of lower channel redundancy illustrated above, and larger weight entropy is beneficial to the model representational capacity~\cite{ding2022re}.

\begin{figure}[t]
  \centering
  \includegraphics[width=.48\textwidth]{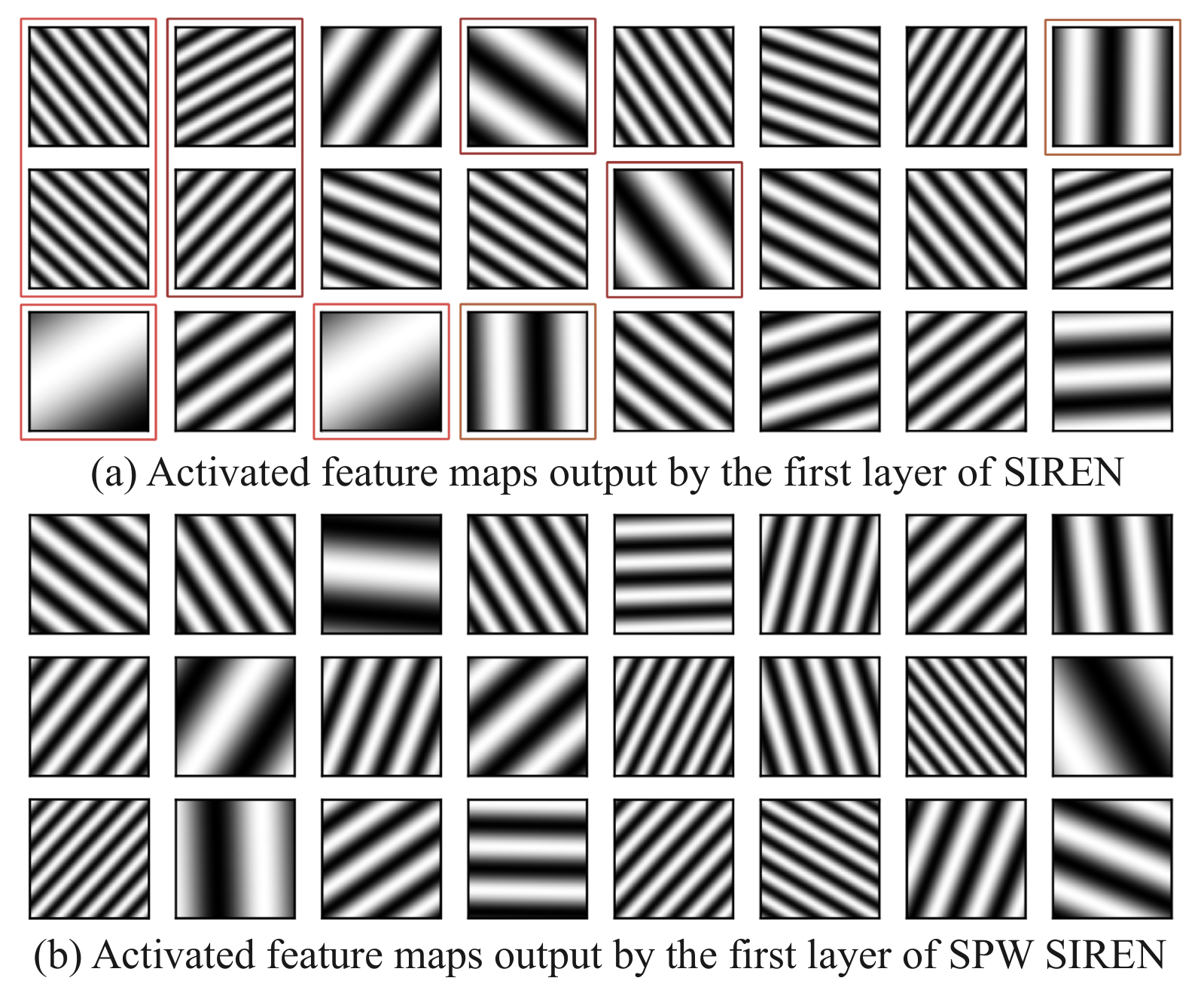} 
  \vspace{-2em} 
 \caption{Activated feature maps output by the first layer of SIREN and SPW SIREN. SIREN learns similar redundant representations as marked by the boxes. While SPW SIREN can learn more distinctive representations.}
 \label{aff}
 \vspace{-1em}
\end{figure}

\vspace{0.5em}
\noindent\textbf{More Novel Representations.} 
Further, we visualize the activated output feature maps of the first layer of SIREN and SPW SIREN.  As exhibited in Fig.~\ref{aff}a, there exists multiple similar activated feature maps output by SIREN, indicating a redundancy of representations. In contrast, the feature maps activated by SPW SIREN exhibit various frequencies (Fig.~\ref{aff}b), indicating that weights with semantic priors can help the model learns more novel representations. This is in alignment with above observations and further validate the effectiveness of SPW.

\vspace{-0.5em}
\subsection{Ablation Study}
\vspace{-0.5em}
\textbf{Generation of the Semantic Vector.} 
Since SPW intends to encode the semantic priors to the weights of INR through the semantic vector, thus the generation of the semantic vector is vital for SPW.
We conduct ablation study that generating the semantic vectors with the output of different SNN stages.
The model tested is SPW SIREN with the architecture of [10$\times$28] trained on Kodak dataset.
As exhibited in Table.~\ref{t2}, the high-level semantic features (namely, features generated by deeper stages) takes the dominating position in influencing the SPW performance that utilizing high-level semantic features along also leads to satisfying results. While using low-level semantic features along leads to largely degraded performance. Furthermore, when combining low-level with high-level semantic features, SPW achieves the best performance. This validates the claim that both low-level and high-level semantic features are important for the enhancement of INR representational capacity. 

\vspace{0.5em}
\noindent\textbf{Search Space of Weight Generation Network.}
The search space of WGN (namely, the size of WGN) is another key point in SPW, which determines the utilization of the semantic vector and the generation of the INR weights.
We conduct ablation study that generate the semantic prior weights by WGNs with different depth and width, shown in Table.~\ref{t3}. 
As can be concluded, the width of WGN influences the effectiveness of SPW slightly. Furthermore, when WGN has fewer than 3 layers, an increase in the depth leads to an enhancement in the performance of SPW. 
However, upon reaching a depth of 3 layers, the performance of SPW peaks and subsequently declines, for that a WGN with 4 layers exhibits slightly inferior performance compared to its 3-layer counterpart.
\begin{table}[t]
\centering
\begin{tabular}{l|cc}
\hline
Selected Stages    & Length of $v_s$& PSNR \\
\hline
Stage 1,2,3 & 192 & 15.32 \\
Stage 1,2,3,4,5 & 576 & 17.21 \\
Stage 1,2,3,4,5,6,7 & 1600 & 26.61\\
Stage 4,5,6,7 & 1408 & 26.24\\
Stage 6,7 & 1024 & 26.03\\
\hline
\end{tabular}
\caption{Results of semantic vectors generated by different stages of EfficientNet.}
\label{t2}
\end{table}

\begin{table}[t]
\centering
\begin{tabular}{l|cccc}
\hline
Depth    & 1 & 2 & 2 & 2 \\
Width    & C & C,C & 4C,C & 8C,C\\
\hline
PSNR     & 22.22 & 25.47 & 25.53 & 25.56 \\
\hline
Depth    & 3 & 3 & 3 & 4 \\
Width    & C,C,C & 4C,4C,C & 8C,8C,C & C,C,C\\
\hline
PSNR     & 26.41 & 26.61 & 26.52 & 26.13 \\
\hline
\end{tabular}
\caption{Results of weight generation network with different depth and width. Depth is the number of fully-connected layers in WGN, width is the number of the output channels in each layer. C is the number of parameters that each layer of INR model required.}
\label{t3}
\vspace{-1em}
\end{table}

\vspace{-1em}
\section{Conclusion}
\vspace{-0.5em}
This paper proposes SPW, the first method that makes INR contain semantic information implicitly through encoding the semantic priors to the weights of INR. 
Experimental results show that SPW can be applied to various INR models and enhance their representational capacity significantly without introducing extra application costs.
Further experiments illustrate that SPW can reduce the weight redundancy and make INR model learn more novel representations, which validates the effectiveness of SPW. 


{
\small

}

\end{document}